\documentclass[runningheads]{llncs}
\usepackage{graphicx}
\usepackage{amsmath,amssymb,bm}
\usepackage{mathtools}

\usepackage[linesnumbered, ruled, vlined]{algorithm2e}

\usepackage{xcolor}
\usepackage{tikz}
\usetikzlibrary{calc, shapes, fit}
\usepackage{pgfplots}
\pgfplotsset{compat=1.10}

\definecolor{tucol1}{rgb}{1.0 0.5 0.05}
\definecolor{tucol2}{rgb}{0.52,0.72,0.10}
\definecolor{tucol3}{rgb}{0.0 0.0 0.8}
\definecolor{tucol5}{rgb}{0.8 0.8 0}
\definecolor{tucol4}{rgb}{0.8 0 0.8}
\definecolor{tucol6}{rgb}{0 0.8 0.8}

\definecolor{tucol-kraeftig1}{RGB}{132,184,25}
\definecolor{tucol-kraeftig2}{RGB}{216,148,39}
\definecolor{tucol-kraeftig3}{RGB}{27,161,175}
\definecolor{tucol-kraeftig4}{RGB}{168,0,135}
\definecolor{tucol-kraeftig5}{RGB}{239,228,32}
\definecolor{tucol-kraeftig6}{RGB}{202,116,40}

\definecolor{tucol-kuehl1}{RGB}{132,184,25}
\definecolor{tucol-kuehl2}{RGB}{233,238,168}
\definecolor{tucol-kuehl3}{RGB}{217,233,229}
\definecolor{tucol-kuehl4}{RGB}{191,218,187}
\definecolor{tucol-kuehl5}{RGB}{113,176,96}
\definecolor{tucol-kuehl6}{RGB}{211,223,99}

\definecolor{tucol-warm1}{RGB}{132,184,25}
\definecolor{tucol-warm2}{RGB}{112,61,46}
\definecolor{tucol-warm3}{RGB}{193,178,40}
\definecolor{tucol-warm4}{RGB}{228,200,38}
\definecolor{tucol-warm5}{RGB}{178,175,132}
\definecolor{tucol-warm6}{RGB}{165,134,83}

\definecolor{tucol-rot1}{RGB}{132,184,25}
\definecolor{tucol-rot2}{RGB}{196,21,58}
\definecolor{tucol-rot3}{RGB}{157,19,42}
\definecolor{tucol-rot4}{RGB}{119,24,40}
\definecolor{tucol-rot5}{RGB}{97,126,31}
\definecolor{tucol-rot6}{RGB}{58,82,11}

\definecolor{tucol-gruen1}{RGB}{132,184,25}
\definecolor{tucol-gruen2}{RGB}{214,223,42}
\definecolor{tucol-gruen3}{RGB}{97,134,39}
\definecolor{tucol-gruen4}{RGB}{148,164,33}
\definecolor{tucol-gruen5}{RGB}{182,201,48}
\definecolor{tucol-gruen6}{RGB}{115,158,64}

\definecolor{tucol-blau1}{RGB}{132,184,25}
\definecolor{tucol-blau2}{RGB}{11,161,226}
\definecolor{tucol-blau3}{RGB}{36,123,196}
\definecolor{tucol-blau4}{RGB}{40,140,141}
\definecolor{tucol-blau5}{RGB}{177,213,230}
\definecolor{tucol-blau6}{RGB}{13,75,127}

\usepackage{hyphenat}
\usepackage{tabularx}
\usepackage{multirow}
\usepackage{makecell}
\usepackage{booktabs}
\usepackage{cite}
\newcolumntype{L}[1]{>{\hsize=#1\hsize\raggedright\arraybackslash}X}
\newcolumntype{R}[1]{>{\hsize=#1\hsize\raggedleft\arraybackslash}X}
\newcolumntype{C}[1]{>{\hsize=#1\hsize\centering\arraybackslash}X}

\usepackage[bookmarks=False,
colorlinks,
urlcolor=blue,
citecolor=black!50!green,
linkcolor=blue
]{hyperref}

\usepackage{cleveref}

\DeclareMathOperator*{\argmin}{argmin}

\newcommand{\ie}{i.e.\xspace}

\newcommand{\wrt}{w.r.t.\xspace}

\newcommand{\qbe}{\textit{query-by-example}\xspace}
\newcommand{\qbs}{\textit{query-by-string}\xspace}

\newcommand{\segfree}{\textit{segmentation-free}\xspace}
\newcommand{\segbased}{\textit{segmentation-based}\xspace}
\newcommand{\Segbased}{\textit{Segmentation-based}\xspace}
\newcommand{\annfree}{\textit{annotation-free}\xspace}
\newcommand{\annbased}{\textit{annotation-based}\xspace}
\newcommand{\lfree}{\textit{learning-free}\xspace}
\newcommand{\lbased}{\textit{learning-based}\xspace}

\newcommand{\pl}{pseudo-label\xspace}
\newcommand{\pls}{pseudo-labels\xspace}

\colorlet{InColor}{tucol-blau2}
\colorlet{OutColor}{tucol-rot2}
\colorlet{TrainColor}{tucol-gruen6}

\tikzset{
	rect/.style={rectangle, rounded corners=2mm, align=center, inner sep=0pt},
	block/.style n args={3}{rect, minimum width=#1cm, minimum height=#2cm, draw=#3, line width=0.2mm},
}

\begin{document}
\title{Annotation-free Learning of Deep Representations for Word Spotting using Synthetic Data
and Self Labeling}
\titlerunning{Annotation-free Learning of Deep Representation for Word Spotting}
%
\author{Fabian Wolf\orcidID{0000-0001-8842-3718} \and
Gernot A. Fink\orcidID{0000-0002-7446-7813}}
\authorrunning{F. Wolf and G. A. Fink}
%
\institute{TU Dortmund University,\\ Department of Computer Science, \\ 44227 Dortmund, Germany\\
\email{\{firstname.lastname\}@cs.tu-dortmund.de}}
\maketitle              
\begin{abstract}
Word spotting is a popular tool for supporting the first exploration of historic, handwritten document collections.
Today, the best performing methods rely on machine learning techniques, which require a high amount of annotated training material.
As training data is usually not available in the application scenario, \annfree methods aim at solving the retrieval task without representative training samples.
In this work, we present an \annfree method that still employs machine learning techniques and therefore outperforms other \annfree approaches.
The weakly supervised training scheme relies on a lexicon, that does not need to precisely fit the dataset.
In combination with a confidence based selection of pseudo-labeled training samples, we achieve state-of-the-art \qbe performances.
Furthermore, our method allows to perform \qbs, which is usually not the case for other \annfree methods.


\end{abstract}

\section{Introduction}
The digitization of documents sparked the creation of huge digital document collections that are a massive source of knowledge.
Especially, historic and handwritten documents are of high interest for historians.
Nonetheless, information retrieval from huge document collections is still cumbersome.
Basic functionalities such as an automatic search for word occurrences are extremely challenging, due to the high visual variability of handwriting and degradation effects.
Traditional approaches like \textit{optical character recognition} often struggle when it comes to historic collections.
In these cases, word spotting methods that do not aim at transcribing the entire document offer a viable alternative \cite{Giotis17}.
Word spotting describes the retrieval task of finding the most probable occurrences of a word of interest in a document collection.
As the system provides a ranked list of alternatives, it is up to the expert and his domain knowledge to decide which entities are finally relevant.

Considering document analysis research, machine learning strongly influenced word spotting methods and a multitude of systems emerged \cite{Giotis17}.
Common taxonomies distinguish methods based on the type of query representation, a previously or simultaneously performed segmentation step and the necessity of a training procedure.
Most systems represent the query either by an exemplar image (\qbe) \cite{Rath07, Retsinas19, Rothacker19} or a string representation (\qbs) \cite{Wilkinson16, Sudholt18, Krishnan19}.
In order to localize a query, the documents need to be segmented into word images.
\Segbased methods \cite{Sudholt18, Retsinas19, Krishnan19} assume that this segmentation step is performed independently beforehand.
In contrast, \segfree methods such as \cite{Rothacker17, Wilkinson17} aim at solving the retrieval and segmentation problem jointly. 
Another distinction commonly made concerns the use of machine learning methods.
So called \lfree techniques rely on expert designed feature representations \cite{Retsinas19, Vats19} and they are usually directly applicable as they do not rely on a learning phase. 
Motivated by the success in other computer vision tasks, machine \lbased techniques and especially convolutional neural networks dominate the field of word spotting today \cite{Sudholt18, Toselli19, Wilkinson17}.

The distinction between \lfree and \lbased word spotting methods suggests that applying machine learning methods is a disadvantage in itself.
This is only true for supervised learning approaches that require huge amounts of annotated training material in order to be successful.
In cases where learning can be applied without such a requirement, we can not see any disadvantage of leveraging the power of machine learning for estimating models of handwriting for word-spotting purposes.
We therefore suggest to distinguish methods based on the requirement of training data.
Methods that do not rely on any manually labeled samples will be termed \annfree as opposed to \annbased techniques relying on supervised learning, as most current word spotting approaches based on deep learning do
\cite{Sudholt18, Toselli19}.
Today, almost all \annfree methods are also \lfree as it is not straight forward to devise a successful learning method that can be applied if manual annotations are not available.
These \lfree methods provide a feature embedding that encodes the visual appearance of a word \cite{Retsinas19, Vats19}.
As no model for the appearance of handwriting is learned, \qbs is usually out of scope for these approaches.

\begin{figure}
	\caption{Semi-supervised training scheme: First an initial model is trained on synthetic data. The model is then iteratively transferred to the target domain by training on confidently estimated samples which are pseudo-labeled with lexicon based recognition.}
	\label{fig:overview}
	\centering
	\input{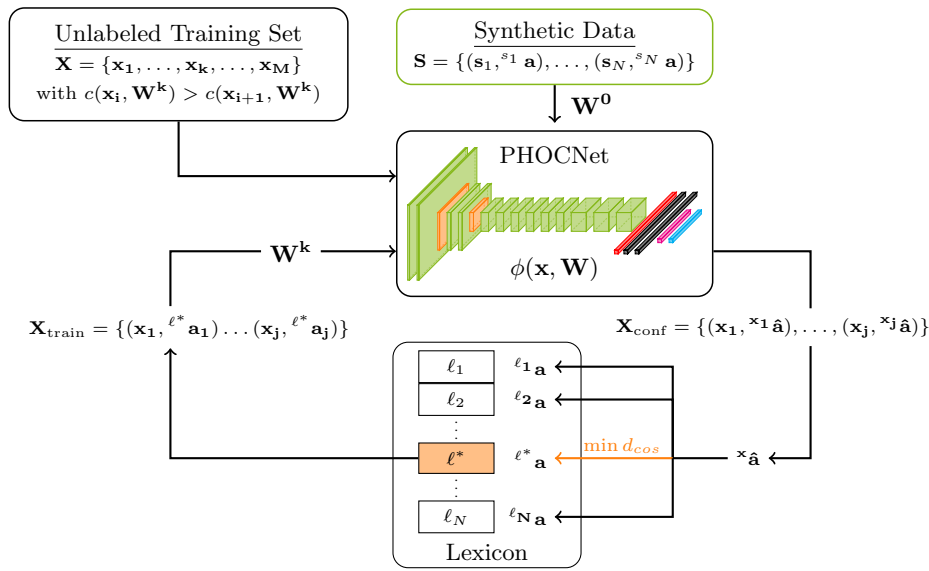}
\end{figure}

In this work, we propose an \annfree method for \segbased word spotting that overcomes this drawback by performing learning without requiring any manually labeled data.
The proposed method uses a synthetic dataset to train an initial model.
Due to the supervised training on the synthetic dataset, the model is capable to perform \qbs word spotting.
This initial model is then transferred to the target domain iteratively in a semi-supervised manner.
Our method exploits the use of a lexicon which is used to perform word recognition to generate \pls for the target domain.
The selection of \pls used to train the network is based on a confidence measure.
We show that a confidence based selection is superior to randomly selecting training samples and already a rough estimate of the lexicon is sufficient to outperform other \annfree methods.
The proposed training scheme is summarized in \autoref{fig:overview}.

\section{Related Work}
In order to solve the retrieval task of word spotting, a system evaluates the similarity between document image regions and a query.
Similar to popular recognition models, many methods exploit the sequential structure of handwriting.
In an early work on \segbased word spotting, Rath and Manmatha proposed to use dynamic time warping to quantify the similarity of two word images based on the optimal alignment of their sequences \cite{Rath07}.
Other sequential models such as \textit{hidden Markov models} (HMM) \cite{Rothacker19, Vidal15} and \textit{recurrent neural networks} \cite{Puigcerver18} were also successfully used for word spotting and they are still popular today
\cite{Toselli19}. 

Traditional feature extraction methods also saw high popularity, due to their success in other computer vision tasks.
In this case, the general approach is to embed the visual appearance of a word image in a feature vector.
This \textit{holistic} representation can then be easily compared to other document regions or image queries with a simple distance measure.
Traditional descriptors based on gradient orientations such as HOG, LBP and SIFT have been shown to be suitable to capture the characteristics of handwriting \cite{Kovalchuk14, Retsinas19, Almazan12}.
Usually the descriptors of an image patch are accumulated in a histogram following the idea of a \textit{Bag of Visual Words (BoVW)}.
Since such a histogram vector neglects any spatial relations between descriptors, it is necessary to combine the approach with an additional model.
For example, \cite{Aldavert15} and \cite{Rusinol15} use a pyramidal scheme to add spatial information, while \cite{Rothacker19} encodes a sequence of BoVW vectors with an HMM. 
As these feature-based approaches only embed the visual appearance, they often struggle when spotting is performed across multiple writing styles.
Queries need to be given in the image domain, which only allows \qbe word spotting.

These limitations motivated the use of \lbased approaches.
In an influential work \cite{Almazan14}, Almazan et al. proposed the use of attribute representations.
The method aims at learning the mapping between word images and a \textit{Pyramidal Histogram of Characters}, which is a binary vector encoding the spatial occurrences of characters.
As the derivation of a PHOC vector from a string is trivial, word images and strings can be mapped in a common embedding space, allowing \qbs.
In \cite{Almazan14}, the visual appearance of a word image is first encoded in a Fisher Vector, followed by a set of Support Vector Machines predicting the presence or absence of each attribute.

The application of neural networks and deep learning also resulted in a strong performance gain in the field of word spotting.
In \cite{Sudholt16}, a convolutional neural network is employed to learn attribute representations similar to \cite{Almazan14}.
Other methods such as \cite{Krishnan16} or\cite{Wilkinson16} also employed neural networks to learn different feature embeddings.
Essentially, the proposed methods based on neural networks significantly outperformed all previous approaches for \qbe and \qbs.
While the discussed networks are usually trained on segmented word images, it has been shown that the approaches can be effectively adapted to the \segfree scenario by using word hypotheses \cite{Rothacker17} or region proposal networks \cite{Wilkinson17}.

Although \lbased methods showed exceptional performances on almost all benchmarks, they rely on a tremendous amount of training data.
Considering the application of a word spotting tool, which is the exploration of a so far unknown document collection, the assumption that annotated documents are available rarely holds.
This problem is far from being exclusive to word spotting and it has been of interest to the computer vision and machine learning communities in general.
\textit{Semi-supervised learning} describes the concept of using unlabeled data in combination with only a limited amount of annotated training samples \cite{Engelen19}.
In \cite{Frinken12}, this approach was successfully applied for a document analysis tasks.
The authors used a word spotting system on a unlabeled dataset to generate additional labeled samples for a handwriting recognition system.
A special type of \textit{semi-supervised} methods employ so called self-labeling techniques \cite{Triguero15}, which have been also studied for neural networks \cite{Lee13}.
In general, an initial model is trained on an annotated dataset and later used to generate labels for unlabeled data of the target domain.
These pseudo-labeled data samples are integrated in the training scheme to further adapt the model.

Transfer-learning describes another approach to reduce the need of training data.
It has been shown that data from another domain can be used to efficiently pre-train a model.
In \cite{Krishnan19}, a synthetic dataset for training word spotting models is proposed.
Annotated training samples are rendered from computer fonts that resemble handwriting.
The resulting dataset is used for pre-training a network that is then fine tuned on samples from the target domain.
As shown in \cite{Gurjar18}, training a model exclusively on synthetic data does not allow for state of the art performances.
Anyways, the amount of training data necessary to achieve competitive results can be reduced significantly.

The lack of training data is a crucial problem for word spotting on historic datasets.
While being clearly outperformed by fully-supervised methods and missing the possibility to perform \qbs word spotting, \annfree, feature-based methods are still receiving attention from the research community \cite{Retsinas19, Vats19}.

\section{Method}
\label{sec:method}
Our method evolves around a basic word spotting system based on an attribute CNN.
We use the TPP-PHOCNet architecture proposed in \cite{Sudholt18} to estimate the attribute representation of an input word image.
A 4-level PHOC representation of partitions 1,2,4 and 8 serves as a word string embedding.
In all experiments, the assumed alphabet is the Latin alphabet plus digits, which results in an attribute vector $\mathbf{a} \in (0,1)^{D}$ with $D=540$.
Given a trained network the system allows to perform word spotting and lexicon-based word recognition, as described in \autoref{sec:spotting}.

The proposed training scheme presented in \autoref{sec:training} does not require any manually annotated training material.
Starting with an initial model, the system exploits the use of automatically generated \pls for the target domain.
In order to enhance the accuracy of the generated labels, only a subset of the predicted \pls is used during the next training cycle.
The selection of samples is based on an estimate of how confident the network is in its predictions.
In this work, we compare the three confidence measures described in \autoref{sec:confidence}.

\subsection{Word Spotting and Recognition}
\label{sec:spotting}
Given a trained PHOCNet with weights $\mathbf{W}$, the network constitutes a function $\phi$ that estimates the desired attribute representation $\mathbf{\hat{a}} = \phi(\mathbf{x}, \mathbf{W})$ for an input word image $\mathbf{x}$.
In the segmentation-based scenario, word spotting is then performed by ranking all word images of the document according to their similarity to a query.
In this work, similarity is measured by the cosine dissimilarity $d_{cos}$ between the estimated attribute vector and the query vector.
Depending on the query paradigm, the query vector  $^q\mathbf{a}$ is either directly derived from a query string or estimated from a query word image $\mathbf{q}$ with $^q\mathbf{\hat{a}} = \phi(\mathbf{q}, \mathbf{W})$.

Word recognition is performed in a similar manner.
Let $\mathbb{L}$ be a lexicon of size $N$ with a set of corresponding attribute representations $^{\ell_i}\mathbf{a}$, $i \in {1, \dots, N}$. 
Based on the estimated attribute vector $^{\mathbf{x}}\mathbf{\hat{a}}$ for the input image $\mathbf{x}$, word recognition reduces to a nearest neighbour search over the lexicon.
Therefore, the recognition result $\ell^{*}$ is given by:
\begin{equation}
	\ell^{*} = \argmin_{\ell \in \mathbb{L}} d_{cos}\left(^{\ell}\mathbf{a}, \mathbf{^x\hat{a}}\right).
\end{equation}

\subsection{Training Scheme}
\label{sec:training}

\begin{algorithm}[t]
\label{alg:training}
\SetKwInOut{Input}{Input}\SetKwInOut{Output}{output}
\caption{Semi-supervised training procedure}
	\Input{Synthetic data $\mathbf{S}=\{(\mathbf{s}_1,^{s_1}\mathbf{a}), \dots, (\mathbf{s}_N,^{s_{N}}\mathbf{a})\}$;
		unlabeled training images $\mathbf{X}=\{\mathbf{x}_1, \dots, \mathbf{x}_M\}$; number of training cycles $K$; \\ PHOCNet $\phi(\cdot,\mathbf{W})$; confidence measure $c(\cdot, \mathbf{W})$;}
	\BlankLine
	Train initial model $\phi(\cdot,\mathbf{W}^0)$ on $\mathbf{S}$\;
	\For{$k\leftarrow 0$ \KwTo $K$}
	{
		Estimate attribute representation $^{\mathbf{x}}\mathbf{\hat{a}} = \phi(\mathbf{x}, \mathbf{W}^k)$ for each element $\mathbf{x}$ in $\mathbf{X}$\;
		Sort $\mathbf{X}$ \wrt confidence: $\mathbf{X}=\{\mathbf{x}_1, \dots, \mathbf{x}_j, \dots, \mathbf{x}_M\}$ with $c(\mathbf{x}_i,\mathbf{W}^k) > c(\mathbf{x}_{i+1},\mathbf{W}^k)$\;
		Select $j$ most confident samples $\mathbf{X}_{conf}=\{\mathbf{x}_1, \dots, \mathbf{x}_j\}$\;
		Generate pseudo labels with word recognition $\mathbf{X}_{train}=\{(\mathbf{x}_1,^{\ell^{*}_1}\mathbf{a}), \dots, (\mathbf{x}_j,^{\ell^{*}_{j}}\mathbf{a})\}$\;
		$\mathbf{W}^{k+1} \leftarrow$ train $\phi(\cdot,\mathbf{W}^k)$ on $\mathbf{X}_{train}$  	
	}
\end{algorithm}

The proposed training scheme is summarized in algorithm \autoref{alg:training}.
Initially, we train the model $\phi(\cdot,\mathbf{W})$ on a purely synthetically generated dataset.
Its images $\mathbf{s}_i$ are generated based on computer fonts that resemble handwriting.
The corresponding attribute representations $^{\mathbf{s}_i}\mathbf{a}$ are known and their creation does not cause any manual annotation effort.

In order to further improve the initial model $\phi(\cdot,\mathbf{W}^0)$, we exploit the unlabeled target dataset.
First, an estimate of the attribute representation $^{\mathbf{x}}\mathbf{\hat{a}}$ is computed for each word image $\mathbf{x}$ in the target dataset $\mathbf{X}$.
As shown in \cite{Gurjar18}, a model only trained on synthetic data does not yield a very high performance and it is likely that the estimated attribute vectors are highly inaccurate.
In order to still derive a reliable \pl, unreliable samples are removed and a lexicon serves as an additional source of domain information.

Inaccurate estimates of attribute vectors are identified by the use of a confidence measure.
Essentially, a confidence measure constitutes a function $c(\cdot,\mathbf{W})$ that quantifies the quality of the network outputs based on its current weights $\mathbf{W}$.
In this work, we investigate three different approaches to measure the network's confidence, see \autoref{sec:confidence}.

In each cycle a fixed percentage of confidently estimated samples is selected.
For each sample in this confident part of the unlabeled dataset $\mathbf{X}_{conf}=\{\mathbf{x}_1, \dots, \mathbf{x}_j\}$ a \pl is generated.
The label $l^{*}$ is derived by performing word recognition with the lexicon $\mathbb{L}$, as described in \autoref{sec:spotting}. 
Therefore, the attribute vector representation used as a training target is derived from the lexicon entry with minimal cosine dissimilarity to the estimated attribute representation $^{\mathbf{x}}\mathbf{\hat{a}}$.
The resulting dataset with \pls is then used for training in order to further adapt the model.

The process of estimating attribute representations based on the current state of the model, selecting confident samples and generating \pls is performed repeatedly for $K$ cycles.
As the training set of pseudo-labeled data $\mathbf{X}_{train}$ is comparably small and the model therefore prone to overfitting the same regularization techniques as proposed in \cite{Sudholt18} are used.
The set of training samples is augmented using random affine transformations.
Based on the predicted labels, word classes are balanced in the resulting augmented training set.
As an additional regularization measure, the PHOCNet architecture employs dropout in its fully connected layers.

\subsection{Confidence Measures}
\label{sec:confidence}
As shown in \cite{Wolf19}, confidence measures allow to quantify the quality of an attribute vector prediction.
Furthermore, recognition accuracies are higher on more confident parts of a dataset, making it a suitable tool for \pl selection.
Following the approach of \cite{Rusakov18}, we model each Attribute $A_i$ as a binary random variable following a Bernoulli distribution.
The output of the attribute CNN is given by $\mathbf{^x\hat{a}}=\phi(\mathbf{x},\mathbf{W})$.
Each element of the output vector is considered an estimate $\mathbf{^x}\hat{a}_i \approx p(A_i=1|\mathbf{x})$ for the probability of the $i$-th attribute being present in the word image $\mathbf{x}$.

\subsubsection{Sigmoid Activation}
Based on the work in \cite{Wolf19}, we derive a confidence measure directly from the network outputs.
Each sigmoid activation provides a pseudo probability for the estimate $\hat{a}_i$.
To estimate the confidence of an entire attribute vector we follow the approach of \cite{Wolf19} and we sum over the estimates of all active attributes.
Neglecting inactive attributes, \ie, attributes with a pseudo probability of $\hat{a_i} < 0.5$, resulted in a slightly better performance in our experiments.
We believe that this is due to the estimated attribute vector being almost binary with only a few attributes close to one.
It seems that the confidence estimation for the high number of absent attributes disturbs the overall assessment of the attribute vector.
The resulting confidence measure $c(\mathbf{x},\mathbf{W})$ is then given by
\begin{equation}
	c(\mathbf{x},\mathbf{W}) = \sum_{\mathbf{^x}\hat{a}_i > 0.5} \phi(\mathbf{x},\mathbf{W})_i \approx \sum_{\mathbf{^x}\hat{a}_i > 0.5} p(A_i=1|\mathbf{x}). 
\end{equation}

\subsubsection{Test Dropout}
Another approach to estimate uncertainty is to use dropout as an approximation \cite{Gal16}.
A confidence measure can be derived by applying dropout layers at test time.
By performing multiple forward passes a variance can be observed for each attribute estimate $\hat{a}_i$.
In this case the assumption is that for a confident prediction the estimate remains constant although neurons are dropped in the dropout layers.
The approach is directly applicable to the PHOCNet as shown in \cite{Wolf19}.
All fully connected layers except the last one are applying dropout with a probability of $0.5$.
We calculate the mean over all attribute variances over $100$ forward passes.
A high confidence corresponds to a low mean attribute variance.

\subsubsection{Entropy}
A well known concept from information theory is to use entropy to measure the amount of information received by observing a random variable \cite{Bishop06}.
The observation of a random variable with minimal entropy does not hold any information.
Therefore, there is no uncertainty about the realization of the random variable.
In this case, a low entropy corresponds to a high confidence in the network's predictions.
Following the interpretation of an attribute as a Bernoulli distributed random variable $A_i$, its entropy is given by
\begin{equation}
	H(A_i) = -\hat{a}_i\log\hat{a}_i - (1-\hat{a}_i)\log(1-\hat{a}_i).
\end{equation}
To model the confidence of an entire attribute vector, we compute the negative joint entropy over all attributes.
As in \cite{Rusakov18}, we assume conditional independence among attributes.
The joint entropy over all attributes is then computed by the sum over the entropies of the individual random variables.
\begin{equation}
	\begin{split}
	c(\mathbf{x},\mathbf{W}) & = - H(A_1, \dots, A_D) =  - \sum_{i=1}^{D}H(A_i) \\
									 & = \sum_{i=1}^{D}\hat{a}_i\log\hat{a}_i + (1-\hat{a}_i)\log(1-\hat{a}_i).
	\end{split}
\end{equation}

\section{Experiments}
We evaluate our method on four benchmark datasets for \segbased word spotting.
In those cases where an annotated training set is available, we do not make use of any labels.
For details on the datasets and the specific evaluation protocols see \autoref{sec:datasets}.
We use \textit{mean average precision} (mAP) in all our experiments to measure performance and to allow for a direct comparison to other methods \cite{Giotis17}.
As the provision of an exact lexicon can be quite a limitation in an application scenario, \autoref{sec:exp_lex} presents experiments on different choices of lexicons.
\autoref{sec:exp_conf} provides an evaluation of different confidence measures and investigates the question whether a confidence based selection of samples is superior to random sampling.
In \autoref{sec:exp_sota}, we compare our method to the state-of-the-art and especially to \annfree methods.

\subsection{Datasets}
\label{sec:datasets}

\subsubsection{George Washington}
The George Washington (GW) dataset has been one of the first datasets used to evaluate \segbased word spotting \cite{Rath07}.
The documents were published by the Library of Congress, Washington DC, USA and they contain letters written by George Washington and his secretaries.
In general, the writing style of the historic dataset is rather homogeneous.
The benchmark contains 4860 segmented and annotated word images.
As no distinctive separation in training and test partition exist, we follow the four-fold cross validation protocol presented in \cite{Almazan14}.
Although we train our network on the training splits, we do not make use of the annotations.
Images from the test split are considered to represent unknown data and are therefore only used for evaluation.

\subsubsection{IAM}
The IAM database was created to train and to evaluate handwriting recognition models \cite{Marti02}.
$657$ different writers contributed to the creation of the benchmark, leading to a huge variety of writing styles.
In total over $115000$ annotated word images are split into a training, validation and test partition.
No writer contributed to more than one partition.
Due to its size and the strong variations in writing styles the IAM database became another widespread benchmark for word spotting \cite{Giotis17}.
The common approach for word spotting is to use each word image (\qbe) or each unique transcription (\qbs) of the test set as a query once.
Stop words are not used as queries but still kept in the test set as distractors.

\subsubsection{Bentham}
The Bentham datasets originated from the project \textit{Transcribe Bentham} and were used for the keyword spotting competitions at the \textit{International Conference on Frontiers in Handwriting Recognition 2014} (BT14)
\cite{Pratikakis14} and at the \textit{International Conference on Document Analysis and Recognition 2015} (BT15) \cite{Puigcerver15}.
The historic datasets contain documents written by the English philosopher Jeremy Bentham and show some considerable variations in writing styles.
Both competitions define a \segbased, \qbe benchmark.
The BT14 set consists of $10370$ segmented word images and a set of $320$ designated queries.
For BT15 the dataset was extended to $13657$ word images and a significantly larger number of queries of $1421$.

\subsubsection{IIIT-HWS}
We use a synthetically generated dataset to train an initial model without any manual annotation effort.
The IIIT-HWS dataset, proposed in \cite{Krishnan19}, was created from computer fonts that resemble handwriting.
Based on a dictionary containing $90000$ words, a total number of $1$ million word images were created and successfully used to train a word spotting model.
We use the published dataset to train our models and did not make any changes to the generation process.

\subsection{Training Details}
As the proposed method is based on the TPP-PHOCNet architecture, we mainly stick to the hyperparameters that have been proven successful in \cite{Sudholt18}.
We train the network in an end to end fashion with Binary Cross Entropy and the ADAM optimizer.
All input word images are inverted, such that the actual writing, presumably dark pixels, is represented by a value of one.
In all experiments, we use a batch size of $10$, weight decay of $5\cdot10^{-5}$ and we employ a momentum with mean $0.9$ and variance $0.999$. 

Our model is initially trained on the IIIT-HWS dataset for $70000$ iterations with a learning rate of $10^{-4}$, followed by another $10000$ training iterations with a learning rate of $10^{-5}$.
We follow the approach of \cite{Gurjar18} and randomly resize the synthetic word images during training to cope with differently sized images in the benchmark datasets. 
Each synthetic word image is scaled by a random factor within the interval $[1,2)$.

The following training phase, which is only weakly supervised by a lexicon, is performed in multiple cycles.
After each cycle, old \pls are neglected and a new set is generated with word recognition and the selection scheme.
On all datasets except IAM we create a total number of $10000$ samples using the augmentation method presented in \cite{Sudholt16}.
Word classes are balanced based on the \pls.
Due to the bigger size of the IAM database, we augment the pseudo-labeled samples to $30000$ images.
For each cycle, the network is trained for one epoch with respect to the augmented training set and a learning rate of $10^{-5}$.
In our experiments, we train the network for $K=20$ cycles.
After selecting $10\%$ of the \pls as training samples during the first $10$ cycles we increase the percentage to $60\%$.

\subsection{Lexicon}
\label{sec:exp_lex}

\begin{table}[t]
\caption{Evaluation of different lexicons. Pseudo-labels are selected randomly in all cases. Results reported as mAP [\%].}
\label{tab:lex}
\begin{tabularx}{\textwidth}{L{1.3} C{0.2} C{0.4}C{0.4} C{0.2} C{0.4}C{0.4} C{0.2} C{0.4}C{0.4} C{0.2} C{0.4}C{0.4}}
\toprule
	\multirow{2}{*}{Lexicon} & & \multicolumn{2}{c}{GW} & & \multicolumn{2}{c}{IAM} & & \multicolumn{2}{c}{BT14} & & \multicolumn{2}{c}{BT15} \\ [0.5ex] 
							& & QbE & QbS & & QbE & QbS & & QbE & QbS & & QbE & QbS\\
	\midrule
							None (*)					& & $46.6$ & $57.9$ & & $16.0$ & $39.5$ & & $18.1$ & - & & $16.4$ & - \\
	\midrule
							Language Based		& & $73.1$ & $64.0$ & & $56.9$ & $77.1$ & & $79.2$ & - & & $65.2$ & - \\
							Closed		 				& & $\bm{87.8}$ & $\bm{87.8}$	& & $\bm{63.7}$	& $\bm{83.6}$ & &       -	& - & &       - & - \\
							Bentham	 					& & 			- & 			- & &		 	- & 			-	& & $\bm{84.3}$ & - & & $\bm{69.1}$ & - \\
\bottomrule
\end{tabularx}
\raggedleft (*) initial model, no weakly supervised training.

\end{table}

In a first set of experiments, we investigate how crucial the prior knowledge on the lexicon is.
All experiments do not make use of a confidence measure but perform the selection of pseudo-labeled samples randomly. 
We experiment with three different types of lexicons and compare the results against the performance of the network after training only on synthetic data.
First we assume that only the language of the document collection is known.
To derive a lexicon for all our datasets, we use the $10000$ most common English words.
Note that this results in $13.5\%$ out-of-vocabulary words on GW, and $10.4\%$ on the IAM database.
Due to the lack of transcriptions, we cannot report out of vocabulary percentages for the Bentham datasets.
As all samples in the GW and IAM datasets are labeled, we can create a closed lexicon containing all training and test transcriptions of the respective datasets.
Even though, this is the most precise lexicon resulting in no out-of-vocabulary words, we argue that in an application scenario an exact lexicon is usually not available.
In case of the Bentham datasets, we investigate another lexicon that is based on the manual line-level annotations published in \cite{Puigcerver15}.
This resembles the case that some related texts, potentially written by the same author, are available and provide a more precise lexicon.

\autoref{tab:lex} presents the resulting spotting performances with respect to the different lexicons.
In general, performances increase substantially by training on the handwritten samples from the target domain under weak supervision.
Already the approximate lexicon based on the modern English language results in high performance gains also for the historic benchmarks.
For the closed as well as the related Bentham lexicon, it can be seen that performances increase with a more precise lexicon.
Nonetheless, we would argue that in the considered scenario only a language based lexicon, which does not require any additional information on the texts besides their language, is a reasonable option.

\subsection{Confidence Measures}
\label{sec:exp_conf}

\begin{table}[t]
\caption{Evaluation of confidence measures. All experiments use a language based lexicon. Results reported as mAP [\%]. Best \annfree results are marked in bold.}
\label{tab:conf}
\begin{tabularx}{\textwidth}{L{1} C{0.2} C{0.4}C{0.4} C{0.2} C{0.4}C{0.4} C{0.2} C{0.4}C{0.4} C{0.2} C{0.4}C{0.4}}
\toprule
		\multirow{2}{*}{Confidence} & & \multicolumn{2}{c}{GW} & & \multicolumn{2}{c}{IAM} & & \multicolumn{2}{c}{BT14} & & \multicolumn{2}{c}{BT15} \\ [0.5ex] 
							& & QbE & QbS & & QbE & QbS & & QbE & QbS & & QbE & QbS\\
	\midrule
							Random 				& & $73.1$ & $64.0$ & & $56.9$ & $77.1$ & & $79.2$ & - & & $65.2$ & - \\
							Entropy			 	& & $79.5$ & $82.1$ & & $\bm{62.6}$ & $\bm{81.3}$ &  & $84.2$ & - & & $75.2$ & - \\
							Sigmoid			 	& & $\bm{83.2}$ & $\bm{82.3}$ & & $62.6$ & $81.0$ &  & $\bm{87.2}$ & - & & $\bm{76.3}$ & - \\
							Test Dropout 	& & $50.4$ & $39.7$ & & $19.5$ & $34.3$ & & $23.4$ & - & & $18.6$ & - \\
	\midrule
		$d_{\textrm{cos}}(\mathbf{^x\hat{a}},\mathbf{^ta})$ 	& & $93.8$ & $94.3$ & & $75.0$ & $87.7$ & & - & - &  & - & -\\
\bottomrule
\end{tabularx}
\end{table}

As discussed in \autoref{sec:confidence}, a confidence measure can be used to identify parts of a dataset that have higher recognition accuracies.
In our experiments, we use the three approaches described in \autoref{sec:confidence} to quantify confidence.
We only select the most confident pseudo-labeled samples to continue training.
Furthermore, we conducted another experiment that uses the cosine dissimilarity between the estimated attribute representation $^{\mathbf{x}}\mathbf{\hat{a}}$ and the actual transcription $^{\mathbf{t}}\mathbf{\hat{a}}$ as a confidence measure.
This is motivated by the idea that a confidence measure essentially quantifies the quality of the attribute estimation, which corresponds to the similarity between estimation and annotation.
Although in practice the cosine dissimilarity cannot be computed without a given annotation, it gives us an upper bound on how well the method would perform with a perfect confidence estimation.

\autoref{tab:conf} presents the results of the experiments, which are conducted with the different confidence measures.
For entropy and sigmoid activations, we observe a performance gain on all benchmarks compared to a random sample selection.
Despite the clear probabilistic interpretation, using entropy performs only on par with sigmoid activations and it is slightly outperformed on the presumably simpler datasets of GW and BT14.
The use of test dropout does not yield any satisfactory results and even performs worse than a random approach.
We observed that test dropout only gives high confidences for rather short words, which makes the selected pseudo-labeled samples not very suitable as training samples.
A longer word potentially provides a bigger set of correct annotation on the attribute level, even in cases where the \pl is wrong.

Considering the use of cosine dissimilarity, it can be seen that a more accurate confidence estimation can still improve performance.
The proposed method in combination with cosine dissimilarity outperforms all other confidence measures, suggesting that the proposed confidence measures are providing suboptimal estimates only.

\subsection{Comparison}
\label{sec:exp_sota}
\begin{table}[t]
\caption{Comparison on GW and IAM. Results reported as mAP [\%]. Best \annfree results are marked in bold, best overall in italic.}
\label{tab:sota_iam_gw}
\begin{tabularx}{\textwidth}{L{1}R{0.6} C{0.4} C{0.3}C{0.3} C{0.4} C{0.3}C{0.3}}
\toprule
	\multirow{2}{*}{Method} & \multirow{2}{5em}{\raggedleft Annotations [n]} & & \multicolumn{2}{c}{GW} & & \multicolumn{2}{c}{IAM} \\ [0.5ex] 
	& & & QbE & QbS & & QbE & QbS\\
\midrule
	Languaged Based \& Sigmoid    & 0 & & $\bm{83.2}$ & $\bm{82.3}$ & & $\bm{62.6}$ & $\bm{81.0}$ \\
\midrule
	Almazan et al. \cite{Almazan12}  & 0 & & $49.4$  & - & & -       & - \\
	Sfikas et al. \cite{Sfikas16}   & 0 & & $58.3$  & - & & $13.2$  & - \\
	DTW 	\cite{Almazan14}	    & 0 & & $60.6$ & - & & $12.3$ & - \\
	FV 	\cite{Almazan14}    	    & 0 & & $62.7$ & - & & $15.6$ & - \\
	Retsinas et al. \cite{Retsinas19} & 0 & & $77.1$ & - & & $28.1$  & - \\
\midrule
	Gurjar et al. \cite{Gurjar18} & 0 	& & $39.8$ & $48.9$ & & $26.2$ & $36.5$ \\
	Gurjar et al. \cite{Gurjar18} & 1000 & & $95.7$ & $96.5$ & & $55.3$ & $74.0$ \\
\midrule
	AttributeSVM \cite{Almazan14} & complete & & $93.0$ & $91.2$ & & $55.7$ & $73.7$\\
	TPP-PHOCNet \cite{Sudholt18} & complete & & $97.9$ & $96.7$ & & $84.8$ & $92.9$\\
	STPP-PHOCNet \cite{Rusakov18} & complete & & $97.7$ & $96.8$ & & $89.2$ & $\mathit{95.4}$ \\
	Deep Embed \cite{Krishnan18} & complete & & $\mathit{98.0}$ & $\mathit{98.8}$ & & $\mathit{90.3}$ & $94.0$\\
	Triplet-CNN \cite{Wilkinson16} & complete & & $98.0$ & $93.6$ & & $81.5$ & $89.4$\\
\bottomrule
\end{tabularx}
\end{table}

\begin{table}[t]
\caption{Comparison for the \annfree, \qbe benchmark on the Bentham datasets. Results reported as mAP [\%]. Best results are marked in bold.}
\label{tab:sota_bt}
\centering
\begin{tabularx}{0.75\textwidth}{L{1}C{0.1}C{0.5}C{0.5}}
\toprule
	\multirow{2}{*}{Method} & & BT14 & BT15 \\ [0.5ex] 
				 & & QbE	& QbE  \\
	\midrule
	Languaged Based \& Sigmoid &						& $\bm{87.2}$ & $\bm{76.3}$ \\
	\midrule
	Aldavert et al. \cite{Aldavert15} 		& & $46.5$ & -		  \\
	Almazan et al. \cite{Almazan14} 			& & $51.3$ & -			\\
	Kovalchuk et al. \cite{Kovalchuk14} 	& & $52.4$ & -			\\
	CVC \cite{Puigcerver15} 							& & -			 & $30.0$ \\
	PRG \cite{Puigcerver15} 							& & - 		 & $42.4$ \\
	\midrule
	Sfikas et al. \cite{Sfikas16} 				& & $53.6$ & $41.5$ \\
	Zagoris et al. \cite{Zagoris17} 			& & $60.0$ & $50.1$ \\
	Retsinas et al. \cite{Retsinas19}			& & $71.1$ & $58.4$ \\
\bottomrule
\end{tabularx}
\end{table}

In order to allow for a fair comparison to the state-of-the-art we only consider the performance of our method with respect to sigmoid activation as a confidence measure and a language based lexicon.
Compared to other \annfree methods, the only additional prior knowledge which we exploit, is the language of the considered documents.
\autoref{tab:sota_iam_gw} reports the performance of the proposed method and other \annfree and \annbased approaches on the GW and IAM dataset.
The best results so far that do not require training material are reported in \cite{Retsinas19}.
Our method achieves higher mean average precisions on both datasets.
Note that the difference is substantially higher in case of the IAM database.
The work in \cite{Retsinas19} is heavily based on a specific feature design to incorporate visual appearance, which is quite suitable for the homogeneous appearance of the GW dataset.
Nonetheless, our method outperforms all other \annfree methods, while the difference is more substantial on datasets as the IAM database where writing styles and visual appearance vary strongly.

In \cite{Gurjar18}, experiments were presented that show how performance increases, when a limited number of annotated samples is used to fine tune a network similar to ours.
While a number of $1\,000$ annotated samples are sufficient to outperform our semi-supervised approach on GW, we still achieve better performances on IAM.
This suggests that our model is able to learn characteristics across different writing styles without relying on any annotations.
Due to the lack of annotated training data from the target domain, our method performs worse compared to fully supervised approaches.

The experiments on both Bentham datasets reported in \autoref{tab:sota_bt} support our observations.
As the benchmarks are considered \annfree, no word image labels are provided.
Therefore, we cannot report any quantitative evaluation of \qbs word spotting.
While outperforming all other methods in the \qbe case, our method additionally offers the possibility to perform \qbs, which is not the case for all other \annfree approaches.

\section{Conclusions}
In this work, we show that an \annfree method for \segbased word spotting, which does not use any manually annotated training material, can still successfully employ machine learning techniques.
Compared to other methods that do not include a learning phase, this leads to significant improvements in performance.
The proposed method relies on a lexicon that provides additional domain information.
Our experiments show that already a language based lexicon, which does not necessarily precisely correspond to the considered documents, is sufficient to achieve state-of-the-art performances.
We successfully make use of a confidence measure to select pseudo-labeled samples during training to boost overall performance.
Additionally, our method provides the capability to perform \qbs word spotting, which is usually not the case for other \annfree approaches.
Therefore, our method is highly suitable for the exploration of heterogeneous datasets where no training material is available.

%
%
%
\bibliographystyle{splncs04}
\bibliography{literature}

\end{document}